\documentclass[11pt]{article}

\usepackage[final]{acl}
\usepackage{multirow}
\usepackage{times}
\usepackage{latexsym}

\usepackage[T1]{fontenc}
\usepackage{amsfonts}

\usepackage[utf8]{inputenc}

\usepackage{microtype}

\usepackage{inconsolata}

\usepackage{graphicx}
\usepackage{booktabs}
\usepackage{amsmath}

\definecolor{revcolor}{RGB}{0,105,140}

\newcommand{\stgcn}{ST-GCN}

\newcommand{\dst}{DSTFormer}
\newcommand{\mpfifty}{MediaPipe-3D}

\newcommand{\mpsixty}{MMPose-2D}

\newcommand{\mpmodel}{\textsc{MediaPipe-3D}}
\newcommand{\unimodel}{\textsc{MMPose-2D}}

\newcommand{\stdv}[1]{{\scriptsize$\pm$#1}}
\newcommand{\fstar}{Fusion$\star$}
\definecolor{pvcyan}{HTML}{23C3D5}
\newcommand{\dashkey}{\textcolor{pvcyan}{%
  \rule[0.42ex]{2.4pt}{1.1pt}\hspace{1.7pt}%
  \rule[0.42ex]{2.4pt}{1.1pt}\hspace{1.7pt}%
  \rule[0.42ex]{2.4pt}{1.1pt}}}

\title{Investigating Multimodal Informativity under Different Partner Visibility Conditions in Video-Mediated Dialogue}

\author{
  Esam Ghaleb\textsuperscript{1}\thanks{\ Equal contribution.}%
  \thanks{\ Corresponding author: \texttt{esam.ghaleb@mpi.nl}} \quad
  Hugh Mee Wong\textsuperscript{2}\footnotemark[1] \quad
  Kristina Kobrock\textsuperscript{3} \\
  \textsuperscript{1}Max Planck Institute for Psycholinguistics \quad
  \textsuperscript{2}Utrecht University \quad
  \textsuperscript{3}Osnabrück University\\
  \texttt{esam.ghaleb@mpi.nl}
}

\begin{document}
\maketitle
\begin{abstract}
Situated language use is multimodal and embodied. For example, gestures can carry information that is absent or underspecified in the speech signal, yet dialogue models typically rely on transcripts alone.
We study how much referential information gestures and their combination with speech carry in multimodal dialogue under different partner visibility conditions.
We build models that identify the intended referent in a video-mediated referential communication game based on either the speech transcript, the skeletal representation of gesture, or both modalities.
Our results show that gesture alone is predictive of the intended referent and that multimodal fusion is most beneficial when the transcript-based model is uncertain. 
In a comparison with human interaction data, we further see pragmatic effects of interlocutor visibility on gesture production and informativeness, as well as an entrainment effect in speech and multimodal, but not gesture, performance across rounds of repeated interaction.
We thus make contributions to the technical modelling of multimodal information in human dialogue and the analysis of human interaction data via trained model representations.
\end{abstract}

\section{Introduction}
Human communication is efficient in part because speakers do not express every aspect of their intended meaning explicitly \citep{grice1975-logic, clark1986-referring}.
Utterances are often vague, ambiguous, or underspecified, requiring listeners to draw on discourse history, shared knowledge, the environment, and assumptions about a speaker's intentions \citep{brennan1996-conceptual, sedivy1999-contextual, hanna2003-common}.

\begin{figure}[t!]
    \centering
    \includegraphics[width=0.99\linewidth]{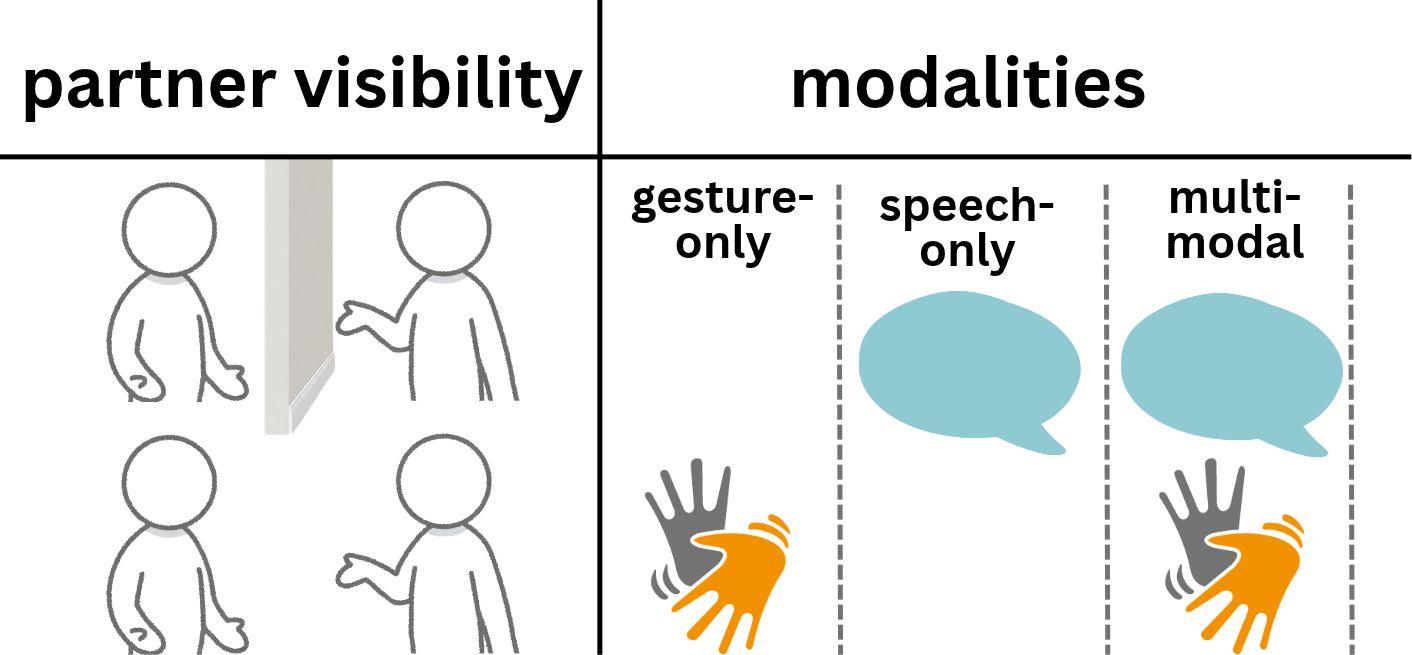}
    \caption{We model and investigate how much information each modality contributes to referential communication when interlocutors can or cannot see each other.}
    \label{fig:1}
\end{figure}

One important source of contextual evidence lies in the multimodal and embodied nature of communication.
For example, in face-to-face interaction, gestures can convey information that is not encoded explicitly in speech and can support reference tracking and the interpretation of ambiguous expressions \citep{beattie1999-iconic, hinnell2020-gesture}.
Iconic gestures are particularly relevant because they depict perceptual or spatial properties of an object, event or action \citep{mcneill1992-hand, kita2003-semantic}.
As such, gestures may provide valuable cues for resolving ambiguity in reference and enhance communication \citep{holle2007-iconic, hinnell2020-gesture, turner_perspective_2026}.
Their communicative use is also shaped by audience design and common ground~\citep{holler_effect_2007}: speakers gesture more when their interlocutor can see them \citep{turner_perspective_2026}. A controlled setup for investigating how much information gesture contributes to dialogue-based referential communication should thus disentangle different modalities and different partner visibility conditions as shown in Figure~\ref{fig:1}.

Reference resolution is also a difficult problem for contemporary language models, as it requires a model to maintain information across the dialogue, recognise when an expression allows for multiple interpretations, and select a referent using contextual evidence.
Evaluations of language models have shown that recognising and disentangling ambiguity remains challenging \citep{liu2023-ambiguity, ma2025-pragmatics}.
Although recent work shows that gesture and dialogue history can improve reference resolution \citep{ghaleb2025-cospeech}, aggregate multimodal performance does not reveal when gesture provides useful information beyond speech.

We propose models to investigate how much referential information computational models can recover from speech and iconic gesture, and when this information complements the verbal signal. Our central aim is not to introduce a new method or architecture, but instead to use existing models to understand how much information each modality contributes to successful referential communication in a multimodal naturalistic dialogue setting.
We analyse models with a video reference game corpus \citep{akamine2025-dataset}, in which interlocutors repeatedly refer to 16 novel objects under different partner-visibility conditions. Given a dialogue utterance, a skeletal gesture sequence, or both modalities, a model must identify the object under discussion. Our contributions are as follows.

\begin{itemize}
    \item We develop gesture-only, speech transcript-only, and multimodal reference-prediction models, comparing complementary 2D and 3D skeletal representations and training-only semantic alignment.\footnote{The code is publicly available at \url{https://github.com/EsamGhaleb/multimodal_informativity/}.}
    
    \item We show that gesture contains recoverable referential information and that its contribution to multimodal prediction is selective: fusion gains are concentrated where the transcript-based model is least certain.

    \item We connect model performance to human interaction patterns, showing that partner visibility affects gesture production and informativeness, while repeated interaction primarily improves transcript-based reference resolution.
\end{itemize}

\section{Related Work}
Research on visually grounded dialogue studies how interlocutors establish and exploit common ground when referring to entities in (partially) shared visual contexts.
The PhotoBook dataset \citep{haber2019-photobook} consists of collaborative dialogues in which participants identify shared images and repeatedly refer to them over the course of an interaction.
Work on PhotoBook shows that dialogue history and previously established referring expressions are important for resolving later references \citep{takmaz2020-refer}.
Related reference games include GuessWhat?! \citep{devries2017-guesswhat}, where a participant identifies an object through a sequence of visually grounded questions, and OneCommon \citep{udagawa2019-onecommon}, where interlocutors must agree on a referent from partially overlapping visual scenes.
These benchmarks, however, represent the interaction primarily through language and visual scene information, without modelling gesture as part of the dialogue.

Research on human multimodal communication shows that co-speech gestures are shaped by speakers' assumptions about their addressees.
Speakers produce more representational gestures when those gestures are visible to a recipient \citep{alibali2001-gesture}, and may produce a greater proportion of nonredundant gestures when concepts are difficult to encode verbally \citep{bavelas2002-gesture}.
In computer-mediated interaction, speakers produce more and larger gestures when they believe that their addressee can see them, while access to cues such as gaze can further modulate gesture production \citep{mol2011-seeing}.
More recently, \citet{turner_perspective_2026} found that speakers in a videoconferencing setting produced more representational gestures when they were visible to the addressee, although simply seeing the addressee's face did not increase gesture production.
These findings motivate our analysis of whether partner visibility affects how much referential information computational models can recover from those gestures.

Particularly relevant to our computational setting, \citet{ghaleb2025-cospeech} study representational co-speech gestures in face-to-face reference games.
Their results indicate that verbally grounded representations capture meaningful properties of gesture more effectively than gesture-only representations and that combining gesture with speech improves referent prediction over speech alone.
In our experiments, we go beyond their work by studying video-mediated dialogue across partner-visibility conditions, by comparing richer skeletal topologies, and by analysing when gesture contributes beyond speech rather than only whether multimodal fusion improves overall prediction.

A related line of work studies embodied reference understanding, where a speaker refers to an object using both language and nonverbal cues such as pointing, pose, or gesture.
YouRefIt \citep{chen2021-yourefit} addresses object localisation from language and pointing gestures, while the Touch-Line Transformer \citep{li2023-touchline} explicitly models the geometric relation between a speaker's pointing direction and the referent.
CAESAR \citep{islam2022-caesar} provides a complementary simulation framework for generating verbal referring expressions and nonverbal cues.
While these studies primarily address spatial localisation and pointing, we investigate iconic gestures produced during collaborative dialogue.

\section{Experimental Setup}

We study reference resolution in a controlled reference-game setting using naturalistic dialogue between human participants.
The task combines a clearly defined prediction problem (identifying an intended referent from a fixed set of candidates) with the variability of spontaneous communication.
Because the objects have no conventional names, participants must develop referring expressions collaboratively and rely on the evolving dialogue context to distinguish the target from competing objects. The game was originally proposed by \citet{eijk2022-cabb} and \citet{rasenberg2022primacy} in the face-to-face interactions CABB corpus.
The resulting interactions therefore provide a suitable setting for investigating whether models can resolve references from linguistic context alone and whether gesture information provides additional evidence.

\subsection{Dataset}
In our evaluations, we use the video-mediated extension of the CABB dataset, which contains interactions between 90 Dutch-speaking participants organised into 45 dyads \citep{akamine2025-cabb-videomediated, akamine2025-dataset}.
Participants were between 18 and 33 years old (mean 22.7), of whom 73 were female and 16 male.
Each dyad completed six rounds, with each round comprising 16 trials.
In each trial, participants interacted with one of 16 novel visual objects, known as \emph{Fribbles} (Appendix~\ref{appendix:all_fribbles}).
Because these objects do not have established labels, participants had to construct and adapt their own descriptions over the course of the interaction.

In each trial, one participant acted as the Director, for whom a target Fribble was marked on their display showing 16 Fribbles.
The Director described the target while the other participant, the Matcher, attempted to identify and select the corresponding object from their own display of Fribbles.
This setup is designed to capture situated communication, where interlocutors collaboratively establish mutually understood referring expressions.

The video-mediated version used in our experiments was collected in a controlled Zoom environment.
The dyads were divided equally among three communication conditions, with 15 dyads assigned to each condition: audio-only (\textsc{ao}), asymmetric video availability (\textsc{asym}), and symmetric video availability (\textsc{sym}). In the audio-only condition, the dyads interacted without any visual signal. In the asymmetric video availability condition, one participant is visible to their partner, while the other participant is not visible. In the symmetric video availability condition, both participants can see each other.
The dataset includes high-quality audio, time-aligned orthographic transcriptions (in Dutch, as well as their English translations), and motion-tracking data.

\subsubsection{Task Definition}
We formulate the task as a referent classification over the set of Fribbles.
Each prediction instance is centred on an annotated gesture interval and pairs one annotated gesture interval with the English translation of an overlapping utterance.
Depending on the input condition, a model receives speech only, gesture only, or both modalities, and must identify the object under discussion.
Because multiple gestures may overlap the same utterance, a single utterance can give rise to more than one gesture-centred instance.

The resulting dataset contains 10,330 unique instances: 2,786 from the audio-only condition, 4,200 from the asymmetric-video condition, and 3,344 from the symmetric-video condition.
An overlapping transcript is available for 10,193 instances.
The output space contains 17 classes: the 16 candidate Fribbles and an additional \textsc{other} class for iconic gesture intervals that do not refer to any of the candidates.
The \textsc{other} class accounts for 238 instances (2.3\%).

\paragraph{Additional training data from face-to-face interactions.}
We additionally used the face-to-face (f2f) corpus from \citet{rasenberg2022primacy} as auxiliary training data only.
This set contains 5,924 gesture instances from 19 participant pairs. Similar to the video-mediated dataset, its RGB and pose sequences were resampled at 29.97 fps using a 500 ms gesture buffer (to cover the gesture unit). Dutch transcripts were translated into English using \texttt{Helsinki-NLP/opus-mt-nl-en}~\cite{tiedemann2020opus}. Overall, the speech model operates on translations of manually or automatically annotated text and does not address automatic speech recognition noise. The possibility of using automatic speech recognition and speech transcription in noisy conditions on the fly can be explored in future work.

\begin{table}[t]
\centering
\scriptsize
\setlength{\tabcolsep}{3.5pt}
\begin{tabular}{@{}p{1.55cm}cp{1.45cm}p{3.15cm}@{}}
\toprule
Model & Nodes & Input & Keypoints \\
\midrule
\mpfifty & 51 & 3D $x,y,z,c$ &
9 body + 21 per hand; MediaPipe world coordinates (illustrated in Figure \ref{fig:mediapipe51_stgcn_3d} in the Appendix). \\
\mpsixty & 69 & 2D $x,y,c$ &
9 body + 21 per hand + 18 face; MMPose (illustrated in Figure \ref{fig:pipeline}). \\
\bottomrule
\end{tabular}
\caption{Gesture skeleton representations.}
\label{tab:pose}
\end{table}

\begin{figure*}
    \centering
    \includegraphics[width=0.9\linewidth]{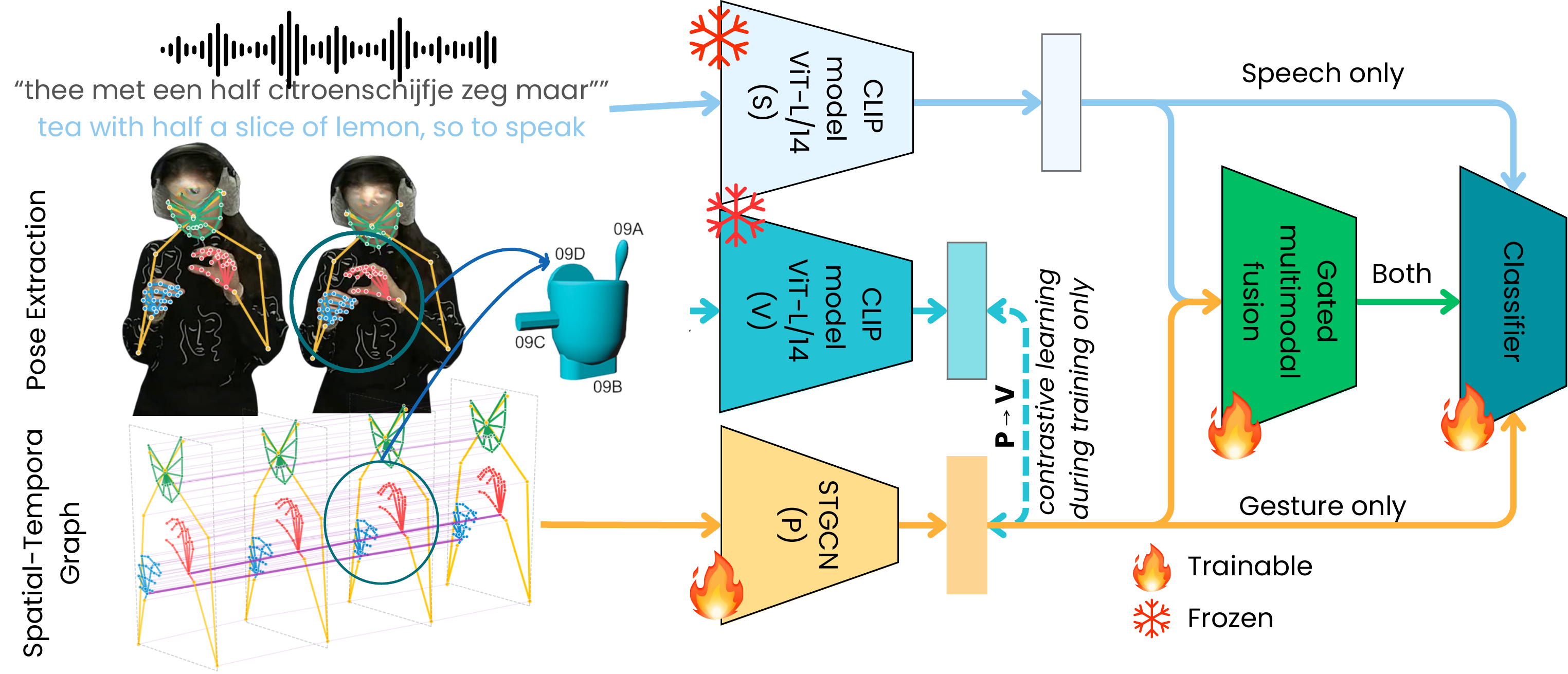}
    \caption{A spatio-temporal graph of skeletal landmarks is encoded by a trainable \stgcn{} (P); the transcript (T) and the referent image (V) are encoded by the same frozen CLIP model. The classifier receives speech only, gesture only, or both through gated fusion. \dashkey{} indicates training-only contrastive alignment $P\!\rightarrow\!V$, not used at inference.}
    \label{fig:pipeline}
\end{figure*}

\subsection{Modelling Approach and Setup}
\subsubsection{Gesture representations}\label{sect:graph_data}
We model gestures as sequences of skeletal configurations from the video frames in the dataset.
We compare two reduced skeletal layouts (Table~\ref{tab:pose}): \mpfifty{}, extracted using MediaPipe Holistic \citep{lugaresi2019-mediapipe}, and \mpsixty{}, extracted using MMPose \citep{mmpose2020}.
The main contrast is that \mpfifty{} provides three-dimensional body and hand coordinates, whereas \mpsixty{} provides two-dimensional coordinates and additionally includes facial landmarks, following the selection used in Uni-Sign \citep{li2025-unisign}.

A video clip of a skeletal sequence is represented as $X \in \mathbb{R}^{C \times T \times V}$, where $C$ is the number of feature channels associated with each landmark, $T$ is the number of video frames, and $V$ is the number of landmarks.
The feature channels contain coordinates and the confidence score provided by the pose estimator.
In addition to joint coordinates and confidence scores, we compute bone features as the offset between each landmark and its parent in the corresponding skeletal topology: $
r_i = X[:, :, i] - X[:, :, c]\textup{,}$ where $c_i$ is the parent of landmark $i$.

We encode the joint and bone streams using separate ST-GCN branches \citep{yan2018-stgcn, jiang2021skeleton} and combine their outputs through learned fusion.
ST-GCN models anatomical connections between landmarks within each frame and temporal connections between corresponding landmarks in adjacent frames.
We also benchmarked DSTFormer \citep{zhang2024-dstformer}, but ST-GCN performed better with both layouts; architecture and benchmarking details are provided in Appendix~\ref{sect:dstformer_results}.
All subsequent analyses therefore use ST-GCN.

\subsubsection{Speech transcript and referent-image representations}
To connect skeletal motion to the semantic content of the task, we derive speech and referent-image representations from a frozen CLIP ViT-L/14-336 model \citep{radford2021-clip}.
Speech is represented using the model's 768-dimensional text projection applied to the aligned English translations of the Dutch utterance.
We apply the corresponding image projection to one normalised rendering of each of the 16 Fribbles, yielding a fixed bank of class prototypes $V_y$.
The text representations provide the input to the speech classifier.
Text and image representations are also used in the frozen teachers for the training-only alignment objectives described in the following subsection.

CLIP was chosen to provide a common frozen text/image space for the controlled alignment study. This does not imply that we consider it to be the most effective text encoder. We have observed that more powerful foundation models, such as Qwen, ignore the additional gesture modality. We suspect this is due to robust text pre-training.

\subsubsection{Classification and multimodal fusion and alignment}
\label{sec:classification_and_alignment_method}
The resulting pipeline of the study is illustrated in Figure~\ref{fig:pipeline}. As illustrated, the speech-only and gesture-only classifiers project their input to 256 dimensions before a two-layer classification head. The fused classifier uses a speech-anchored residual fusion with a sample-specific gate,
\[
    F = W\!\left(s + \sigma\!\left(g_\theta([p; s; |p-s|])\right)\odot p\right)\textup{,}
\]
where $p$ and $s$ are the pose and speech representations. The baseline classification models minimise class-balanced cross-entropy.

In addition to the baseline classification objective, which minimises class-balanced cross-entropy, we optionally add a symmetric multi-positive supervised InfoNCE term:
\begin{equation}
\begin{aligned}
 \mathcal{L}&=\mathcal{L}_{\mathrm{CE}}
 +\lambda\sum_{e\in E}\omega_e\mathcal{L}_{\mathrm{NCE}}^{(e)},\\
 \lambda&\in\{0.1,0.2,0.3\}.
\end{aligned}
\end{equation}
Students and teachers are $\ell_2$-normalized.
For each anchor, every globally gathered representation with the same referent label is a positive and only different-label representations are negatives.
The loss averages student-to-teacher and teacher-to-student directions.
Alignment projectors map learned representations to the frozen 768-dimensional CLIP space; the temperature starts at 0.07,
with a five-epoch warm-up that protects the classifier from an initially dominant auxiliary loss. 

In this work, we already employ multimodal speech representations through CLIP Embeddings. As such, this alignment setup was mainly applied on the gesture (i.e., Pose) and frozen Fribble-image representations (Vision) of the objects: P$\rightarrow$V. The rationale and results of this integration are further explained in Section \ref{sect:alignment_impact}.

\subsubsection{General-purpose VLM baseline}
\label{sect:general_purpose_vlm}
As a task-level reference point, we also evaluate Qwen3-VL \citep{qwen3} under transcript-only, gesture-only, and multimodal input conditions.
Fine-tuning improves transcript-based reference resolution, but adding gesture provides no clear benefit.
Because this baseline jointly entangles gesture perception, temporal modelling, and multimodal integration, we use the controlled skeletal models described above to assess gesture informativity more directly.
Full experimental details and results are reported in Appendix~\ref{appendix:qwen}.

\subsection{Evaluation Protocol and Metrics}\label{sec:protocol}
\paragraph{Data splits.}
We use two five-fold cross-validation protocols for the video-mediated data. In trial-level cross-validation, all gesture-centred instances from the same trial are assigned to one fold. Other trials from the same dyad can occur in training. 
The trial-level evaluation is our primary and main protocol since gestures are dyad- and speaker-specific behaviors. 
For the generalisation test, in dyad-held-out cross-validation, all trials from a dyad are assigned to one fold. This tests whether the models can resolve references made by interlocutors whose dialogue was absent from training and development.
In each protocol, we test on fold $k$, use fold $(k+1)\bmod 5$ for development and train on the other three folds.
Face-to-face data are added to the training folds only and never used for development or testing.

\paragraph{Training.}
Models are trained with AdamW for up to 256 epochs using conservative pose augmentation (Appendix~\ref{app:technical}).
We select checkpoints by development macro-F1 and apply early stopping.

\paragraph{Metrics.}
We report top-$k$ accuracy for $k \in \{1, 2, 3, 4, 5\}$ and macro-F1.
Because the Fribbles share visually similar subparts, we measure ranking quality using the mean top-1-to-top-5 utility
\begin{equation}
U_{1:5}=
\frac{1}{5}\sum_{k=1}^{5}
\mathbb{I}\!\left(r_i\leq k\right)\textup{,}
\label{eq:topk-utility}
\end{equation}
where $r_i$ is the rank assigned to the gold referent.
This equals $1.0$ when the referent is ranked first, $0.8$ when second, and $0$ when it falls
outside the top five.

\begin{table*}[t]
\centering
\small
\setlength{\tabcolsep}{3.6pt}
\begin{tabular}{@{}clllcccccc@{}}
\toprule
CV & Input & Layout & Align. & Top-1 & Macro-F1 & Top-2 & Top-3 & Top-5 & $U_{1:5}$ \\
\midrule
\multirow{9}{*}[-2pt]{\rotatebox[origin=c]{90}{\shortstack{\textit{Trial-level} \\ \textit{(primary setup)}}}}
 & Gesture & MediaPipe-3D & --- & 19.5\stdv{1.1} & 18.2\stdv{1.2} & 32.5\stdv{1.8} & 41.6\stdv{1.6} & 55.5\stdv{1.9} & 39.6\stdv{1.5} \\
 & Gesture & MMPose-2D & --- & 20.4\stdv{1.4} & 18.9\stdv{1.3} & 32.6\stdv{2.2} & 42.1\stdv{2.5} & 56.4\stdv{3.3} & 40.3\stdv{2.4} \\
 & Gesture & MediaPipe-3D & P$\rightarrow$V & 20.4\stdv{1.1} & 18.8\stdv{1.0} & 32.5\stdv{1.7} & 41.6\stdv{2.1} & 55.2\stdv{2.3} & 39.7\stdv{1.9} \\
 & Gesture & MMPose-2D & P$\rightarrow$V & 20.1\stdv{1.0} & 18.6\stdv{1.2} & 32.6\stdv{2.2} & 41.5\stdv{2.4} & 55.0\stdv{2.7} & 39.6\stdv{2.1} \\
\addlinespace[2pt]
 & Speech & --- & --- & 43.8\stdv{3.4} & 42.8\stdv{3.2} & 58.0\stdv{1.8} & 66.5\stdv{2.1} & 76.7\stdv{1.9} & 63.5\stdv{2.1} \\
\addlinespace[2pt]
 & Gesture + speech & MediaPipe-3D & --- & 45.7\stdv{3.8} & 44.3\stdv{3.6} & 60.2\stdv{3.3} & 68.7\stdv{2.6} & 79.2\stdv{2.4} & 65.7\stdv{2.9} \\
 & Gesture + speech & MMPose-2D & --- & 46.0\stdv{2.8} & 44.4\stdv{2.7} & 60.1\stdv{2.0} & 68.2\stdv{2.1} & 78.9\stdv{1.2} & 65.5\stdv{1.9} \\
 & Gesture + speech & MediaPipe-3D & P$\rightarrow$V & 46.1\stdv{4.3} & 44.9\stdv{3.8} & 60.4\stdv{2.4} & 68.8\stdv{1.9} & 79.4\stdv{1.5} & 66.0\stdv{2.2} \\
 & Gesture + speech & MMPose-2D & P$\rightarrow$V & 46.1\stdv{3.4} & 44.4\stdv{3.3} & 61.3\stdv{2.3} & 69.2\stdv{2.0} & 79.8\stdv{0.9} & 66.3\stdv{2.0} \\
\midrule
\multirow{4}{*}[-2pt]{\rotatebox[origin=c]{90}{\itshape\begin{tabular}{@{}c@{}}Dyad-\\held-out\end{tabular}}}
 & Gesture & MMPose-2D & --- & 18.4\stdv{2.8} & 17.0\stdv{2.9} & 30.0\stdv{3.2} & 39.2\stdv{3.1} & 53.1\stdv{2.7} & 37.4\stdv{2.8} \\
\addlinespace[2pt]
 & Speech & --- & --- & 38.5\stdv{1.6} & 37.6\stdv{1.9} & 53.4\stdv{1.4} & 63.0\stdv{1.8} & 74.1\stdv{1.5} & 59.6\stdv{1.2} \\
\addlinespace[2pt]
 & Gesture + speech & MMPose-2D & --- & 39.8\stdv{1.3} & 38.7\stdv{1.3} & 55.6\stdv{1.2} & 64.6\stdv{1.4} & 75.6\stdv{2.4} & 61.3\stdv{1.2} \\
 & Gesture + speech & MMPose-2D & P$\rightarrow$V & 41.1\stdv{1.5} & 39.9\stdv{2.0} & 56.0\stdv{0.8} & 65.5\stdv{2.0} & 76.6\stdv{2.2} & 62.2\stdv{1.5} \\
\midrule
 & \textit{chance} & --- & --- & 5.9 & --- & 11.8 & 17.6 & 29.4 & 17.6 \\
\bottomrule
\end{tabular}
\caption{Mean prediction performance over the five test folds (\stdv{~}: standard deviation across folds), for trial-level and dyad-held-out cross-validation. $U_{1:5}$ is the mean top-1--top-5 utility. $P\!\rightarrow\!V$ is training-only gesture-to-image alignment at $\lambda=0.3$. Every gesture$+$speech row improves reliably over the speech baseline. Evaluated on the instances with an overlapping transcript.}
\label{tab:main_table}
\end{table*}

\section{Informativity of speech, gesture, and their fusion}\label{sec:rq1}
Table~\ref{tab:main_table} summarises unimodal and multimodal performance on held-out trials and held-out dyads.
We first establish how much referential information is recoverable from gesture and speech separately, before asking whether gesture improves prediction when combined with speech.

\subsection{Gesture alone is a genuine referential channel}
At the trial-level cross-validation, the gesture-only ST-GCN models achieve 19.5\% top-1 accuracy with \mpmodel{} and 20.4\% with \unimodel{}, against a chance level of 5.9\%.
Their top-5 accuracies reach 55.5\% and 56.4\%, respectively, placing the correct referent within a five-item shortlist for more than half of the gesture instances.
Gesture therefore provides a substantial referential signal, although it remains ambiguous: the sharp increase from top-1 to top-5 accuracy is consistent with the Fribbles sharing visually similar subparts.

This result is robust across the two skeletal representations.
Despite differences in dimensionality, landmark count, facial coverage, and coordinate space, \mpmodel{} and \unimodel{} differ by less than one percentage point on most reported metrics.
The gesture-informativity result therefore does not depend strongly on either of the two pose layouts we consider.

\subsection{Speech is stronger, but gesture provides complementary information}
At trial-level cross-validation, the speech-only model achieves 43.8\% top-1 and 76.7\% top-5 accuracy.
This comparatively strong performance reflects both the availability of the complete utterance overlapping each gesture interval and the use of pre-trained CLIP representations.

Combining gesture with speech produces a modest but consistent improvement.
Without semantic alignment, MMPose layouts reach up to 46.1\% top-1 accuracy, approximately 2.2 percentage points above the speech-only model.
Fusion also improves accuracy at the remaining top-$k$ thresholds. These aggregate gains are modest, suggesting that gesture provides useful additional evidence for a subset of instances rather than improving reference resolution uniformly. Section~\ref{sec:when_gesture_helps} examines where these gains arise.

\begin{table*}[th]
\centering
\small
\setlength{\tabcolsep}{4pt}
\small
\setlength{\tabcolsep}{4pt}
\begin{tabular}{@{}lcccc@{}}
\toprule
 & \multicolumn{2}{c}{Trial-level CV} & \multicolumn{2}{c}{Dyad-held-out CV} \\
\cmidrule(lr){2-3} \cmidrule(lr){4-5}
Model & Top-1 & $\Delta$ vs speech [95\% CI] & Top-1 & $\Delta$ vs speech [95\% CI] \\
\midrule
Speech only & 43.8\stdv{3.4} & --- & 38.5\stdv{1.6} & --- \\
Gesture only & 20.4\stdv{1.4} & --- & 18.4\stdv{2.8} & --- \\
Gated fusion & 46.0\stdv{2.8} & +2.2 [+1.0, +3.4] & 39.8\stdv{1.3} & +1.3 [$-$0.0, +2.7] \\
Concatenation & 45.8\stdv{3.1} & +2.0 [+0.6, +3.4] & 40.3\stdv{1.2} & +1.8 [+0.9, +2.8] \\
Cross-attention & 46.6\stdv{3.3} & +2.8 [+1.2, +4.3] & 40.4\stdv{1.6} & +2.0 [+0.7, +3.2] \\
Late fusion (trained) & 45.7\stdv{3.5} & +1.9 [+0.6, +3.2] & 40.3\stdv{0.8} & +1.8 [+0.7, +2.8] \\
Gated + $P\!\rightarrow\!V$ (Fusion$\star$) & 46.1\stdv{3.4} & +2.4 [+1.0, +3.7] & 41.1\stdv{1.5} & +2.6 [+1.2, +3.9] \\
\bottomrule
\end{tabular}
\caption{Fusion mechanisms with identical encoders, folds, and settings (MMPose-2D). Top-1 is the mean over folds, $\Delta$ is the paired difference from speech only in percentage points.}
\label{tab:fusion_ablation}
\end{table*}

\subsection{Dyad-independent evaluation.}
For generalisation, we additionally evaluate the models using dyad-level splits, ensuring that neither dyads nor individual participants overlap between the training and test sets. For this evaluation, we use the MMPose-2D layout.
The main findings remain consistent: gesture-only performance remains above chance, multimodal fusion outperforms the speech-only model, and alignment further improves fusion performance. Compared with the trial-level split, gesture-only performance decreases only marginally ($20.4 \rightarrow 18.4$), whereas speech-only
performance exhibits a more substantial decline ($43.8 \rightarrow 38.5$).

This discrepancy suggests that speech-based predictions benefit more
from trial-level splitting, potentially due to the recurrence of
lexical expressions across rounds. Iconic gestures, by contrast,
may generalise better across dyads because they encode properties
of the referent rather than dyad-specific conventions. The best fusion model outperforms the speech-only model
($\Delta = +2.5$).

\subsection{Pose-to-image alignment improves multimodal fusion}
\label{sect:alignment_impact}
We next ask whether training-only semantic alignment can make the gesture representation more useful to the fusion model.
The fusion architecture described in Section~\ref{sec:classification_and_alignment_method} integrates gesture through a gated mechanism into the speech transcript representation.
We now ask whether the two representations can be made more compatible before this integration.
We focus on pose-to-vision alignment ($P\!\rightarrow\!V$), which encourages the learned pose representation to approach the frozen CLIP representation of the target Fribble through a contrastive learning objective.
This alignment objective is only used during training and does not alter the model inputs or test-time computation.

As shown in Table~\ref{tab:main_table}, pose-to-image alignment gives similar performance with both skeletal layouts.
For \unimodel{} at $\lambda=0.3$, top-1 accuracy increases modestly relative to the matched unaligned fusion model. The same alignment objective does not reliably improve gesture-only classification. Top-1 accuracy changes by $+0.96$ points ($p=.130$) for \mpmodel{} and $-0.37$ ($p=.520$) for \unimodel{}.
Thus, $P\!\rightarrow\!V$ alignment does not make the gesture representation more discriminative on its own.

\subsection{Fusion ablation}
Table~\ref{tab:fusion_ablation} compares different fusion mechanisms under trial-level and dyad-held-out cross-validation, while keeping the encoders, folds, and training settings fixed.

Across both evaluation protocols, standard fusion mechanisms differ
by at most 0.9 percentage points and consistently outperform the speech-only model in mean Top-1 accuracy, suggesting that our findings
are not specific to a particular fusion architecture.
Fusion$\star$ performs competitively under both evaluation protocols
and results in the largest improvement over speech alone under
dyad-held-out evaluation ($\Delta = +2.6$ percentage points).
These results support the use of multimodal fusion as an analytical
means of investigating the relative and complementary informativity
of gesture and speech across visibility conditions, rather than as
an architectural contribution in itself.

\subsection{Does the gesture need to match the utterance?}
To assess whether fusion performance depends on the content of both modalities, we conduct mismatch controls under dyad-held-out evaluation. We replace one modality
with an instance from the same fold, dyad, and speaker, but a different trial and referent. Replacing the gesture reduces Top-1 accuracy from 41.5\% to 36.6\% ($\Delta = 4.5$), falling
below the speech-only baseline (38.5\%). This indicates that the benefits of fusion depend on the
gesture content rather than on the inclusion of an
additional encoder. Replacing speech gives a substantially larger performance drop ($\Delta = 35.1$), consistent with the semantic, pragmatic, and temporal interdependence of co-speech gestures and their accompanying utterances.

\emph{For the remaining analyses, we use the \unimodel{} ST-GCN as the gesture-only model at trial-level, as it is our original and primary setup.
For multimodal prediction, we use the \unimodel{} with pose-to-image alignment at $\lambda=0.3$, which we denote \fstar{}.}

\section{When Does Gesture Provide Additional Information?}\label{sec:when_gesture_helps}

Based on the literature and our own analysis of the human interactions in the dataset, we analyse what the pragmatic contribution of multimodal communication is in referential communication.

\subsection{Effects of interlocutor visibility}\label{sec:rq5}
First, we expect the visibility of the interlocutor to play a role in multimodal communication. In line with~\citet{turner_perspective_2026}, we expect gestures to be used more if interlocutors can see each other compared to an audio-only condition. However, when analysing whether a gesture is present in an utterance or not in the dataset, we find that gestures are not used substantially more when interlocutors can see each other compared to when they share audio only (M=0.03, CrI=[-0.16, 0.21], pd=62\%, see Figure~\ref{fig:human_gesture_cond} and Table~\ref{tab:gesture_present_visibility_human} in Appendix~\ref{app:posterior}).

\begin{figure}[t]
    \centering
    \includegraphics[width=0.8\linewidth]{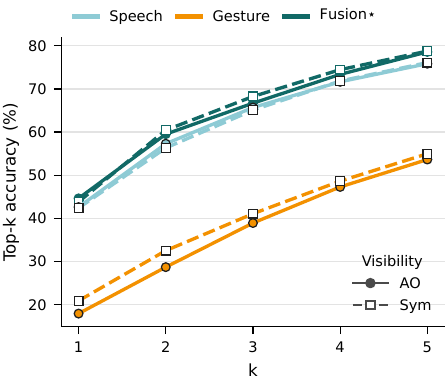}
    \caption{Top-k accuracy by visibility condition and embedding type.}
    \label{fig:top_k_by_condition}
\end{figure}

Being interested in the pragmatic informativity of gestures for multimodal referential communication, we look at the informativity of speech, gesture and their fusion under different visibility conditions that were collected with the dataset. Specifically, we ask: Does the distribution of information in the modalities change depending on whether interlocutors interact in an audio-only setting compared to when they see each other in a video-mediated interaction? \emph{Importantly, this allows us to go beyond the current state-of-the-art when analysing human communication in multimodal interaction by measuring the informativity and contribution of modalities for successful reference resolution}. 
Figure \ref{fig:top_k_by_condition} shows that the performance of the speech and fusion model does not change with visibility conditions. However, for the gesture model, we see a difference between visibility conditions, where the visibility of interlocutors leads to higher classification accuracy, i.e., more information in the gesture representations.

To quantify these differences, we fit a Bayesian binary logistic regression model predicting whether the model made the correct target selection (top-1 accuracy) by visibility condition (\textit{Audio-only}, \textit{Symmetric}), embedding type (\textit{Gesture}, \textit{Speech}, \textit{Fusion*}) and their interaction. We find substantial main effects for condition and embedding type as well as substantial interaction effects (see Table~\ref{tab:posterior_summary_comm} in Appendix~\ref{app:posterior}). We compute posterior difference distributions and report the posterior changes in accuracy in percentage. In particular, while the performance of the speech and fusion model does not depend on visibility conditions (speech: M=0.007, CrI=[-0.03, 0.02], pd=73.10\%, fusion: M=-0.01, CrI=[-0.04, 0.01], pd=85.20\%), the performance of the gesture model improves substantially by about 3\% with visibility (M=0.03, CrI=[0.01, 0.05], pd=99.75\%). 

Gestures produced when no partner can see them are still recoverable by the pose encoder, but they carry less referent information that is usable on top of speech, which is what one would expect if a portion of them are self-oriented rather than communicative.

\begin{figure}[t]
    \centering
    \includegraphics[width=0.9\linewidth]{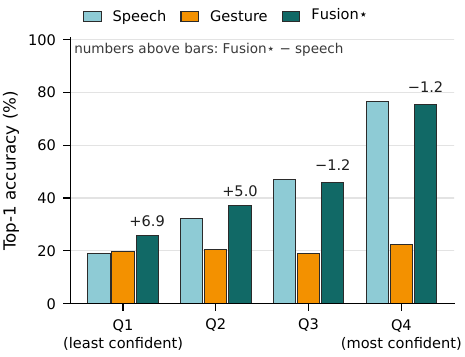}
    \caption{Top-1 accuracy per modality by speech-confidence quartile. Gains from gesture integration concentrate where speech is uncertain.}
    \label{fig:best_uncertainty}
\end{figure}

\subsection{Gestures are especially informative when speech is uncertain}
\label{sec:uncertainty}
We now analyse how gestures modulate speech models' performance and confidence, and we ask whether the gain is organised by speech confidence.
We define speech confidence as the largest probability the speech-only classifier assigns to any referent. Within each protocol, we divide the eligible instances into quartiles of this held-out confidence score and compare \fstar{} with speech on the same instances (Figure~\ref{fig:best_uncertainty}). 

In the least-confident trial-level quartile, speech accuracy is 18.8\% and \fstar{} reaches 25.7\%, a gain of 6.9 points. In the second quartile, accuracy rises from 32.3\% to 37.3\%. The estimates in Q3 and Q4 are $-1.2$ and $-1.2$ points. On held-out dyads, we observe similar patterns. 

\subsection{Entrainment across rounds}\label{sec:rq4}
Second, we look at the contribution of modality-specific information on the lexical entrainment of references across rounds of interactions between dyads. Based on well-known findings from the pragmatics literature on the conventionalization of referring expression over repeated rounds of a referential task~\citep{clark1986-referring}, we expect the same pattern to be present in the dataset where participants referred to the same referents repeatedly over six rounds. We use a Bayesian binary logistic regression model to analyse the probability of a gesture being present in an utterance depending on the round (1-6). We find that the probability of a gesture being present in an utterance decreases with consecutive rounds (M=-021, CrI=[-0.28, -0.154, pd=100\%, see Figure~\ref{fig:round_gesture_human} and Table~\ref{tab:gesture_present_round_human} in Appendix~\ref{app:posterior}).

\begin{figure}[t]
    \centering
    \includegraphics[width=0.9\linewidth]{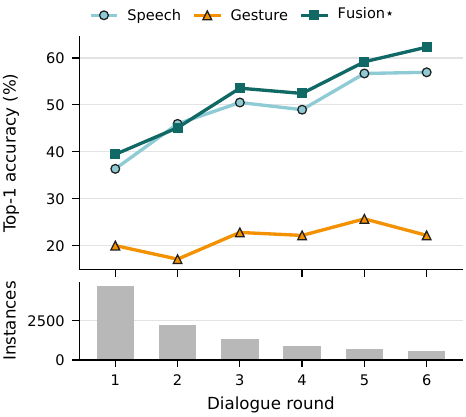}
    \caption{Accuracies of different modalities (lines) and the distribution of gestures (bars) across rounds.}
    \label{fig:round_model}
\end{figure}

Speech-only classification accuracy increases from 36.4\% in round 1 to 56.9\% in round 6 (Figure~\ref{fig:round_model}), consistent with later descriptions becoming easier for the model to resolve. Gesture-only accuracy is flatter: 19.7-22.0\% for \mpmodel{} and 20.0--22.2\% at the endpoints for \unimodel{}. Fusion follows speech.
A Bayesian binary logistic regression model predicting correct classification by embedding type (\textit{Gesture}, \textit{Speech}, \textit{Fusion*}) and round (1-6) shows substantial differences between embedding types (see Table~\ref{tab:gesture_present_round_model} in Appendix~\ref{app:posterior}). More importantly, for gestures, accuracy does not depend on round (M=0.06, CrI=[0.027, 0.088], pd=99.99\%). But looking at speech (M=0.19, CrI=[0.160, 0.212], pd=100.00\%) and fusion models (M=0.20, CrI=[0.170, 0.222], pd=100.00\%), accuracy increases with rounds.
Accordingly, these results are compatible with lexical entrainment in speech but do not identify an independent gesture-entrainment effect.

\section{Discussion and Conclusion}

In conclusion, our models and analyses make important contributions to two sides of the research problem, namely reference resolution in multimodal interaction. First, we built models and trained them to identify the speaker's intended referent based on challenging multimodal, ambiguous, and naturalistic data of human dialogue. Our results suggest that a substantial amount of information on the target referent can be recovered from iconic co-speech gestures alone, corroborating their communicative and informative function in human dialogue. We further showed that including gesture improves speech-only models and that gesture information improves model performance exactly in those speech segments where the model is least certain about how to resolve ambiguity of the speech transcripts. We used pose-to-image alignment to improve multimodal ranking even more.
Second, we connected the model-based analysis of gesture, speech, and multimodal embeddings to partner visibility and entrainment effects seen in the human interaction data. This approach allowed us to investigate the role of co-speech gestures in human language use by quantifying their semantic and pragmatic informativity based on interaction, visual alignment, and the gesture's specific role in dialogue.

\section*{Limitations}
\paragraph{Dataset and task scope.}
Our experiments use a closed-set reference task involving 16 unfamiliar visual objects.
The controlled setting enables direct comparison of modalities, but it is unclear whether the findings generalise to other languages, object domains, dialogue tasks, or less constrained interactions.

\paragraph{Translated transcripts.} The existing CABB studies with human participants were conducted in Dutch. In this study, we focus on English, as the pre-trained language backbones were predominantly trained on English data.
Given the substantial resourced required to collect high-quality human-annotated data from human participants, we have opted to use machine-translated versions of the original Dutch (manually produced) transcripts rather than collect equivalent data in English from scratch.
However, we acknowledge that machine translation can introduce other sources of noise, such as inaccuracies in the translations.

\paragraph{Gesture-centred sampling.}
rPrediction instances are defined around annotated iconic gesture intervals.
The experiments therefore measure the information available once a gesture has already been identified, rather than addressing gesture detection in continuous interaction.
They also exclude non-iconic gestures that may contribute to reference resolution.
Because multiple gestures can overlap the same utterance, some transcript content occurs in more than one instance, although instances from the same trial are kept within the same cross-validation fold.

\paragraph{Asymmetry between modalities.}
The gesture model receives a temporally bounded skeletal sequence, whereas the transcript model receives the complete overlapping utterance.
The comparison consequently reflects the information available under these input definitions rather than an intrinsically matched comparison between modalities.

\section*{Acknowledgements}
We thank the anonymous reviewers for their suggestions, as well as Albert Gatt from Utrecht University for the helpful discussions.
HMW is supported by the Dutch Research Council (NWO) through the AiNed Fellowship Grant \texttt{NGF.1607.22.002}.
Parts of this work were conducted using the Dutch national e-infrastructure with the support of the
SURF Cooperative using grant no. \texttt{EINF-19641}.

\bibliography{refs}

@misc{akamine2025-cabb-videomediated,
  author       = {Akamine, Sho and Dingemanse, Mark and Meyer, Antje and {\"O}zy{\"u}rek, Asli},
  title        = {Video-mediated communication modulates conversational features of speech more than gesture in novel referential communication},
  year         = {2025},
  month        = nov,
  day          = {27},
  publisher    = {PsyArXiv},
  doi          = {10.31234/osf.io/vjwpz_v1},
  url          = {https://doi.org/10.31234/osf.io/vjwpz_v1},
  note         = {Preprint}
}

@misc{akamine2025-dataset,
  author       = {Akamine, Sho and Dingemanse, Mark and Meyer, Antje and {\"O}zy{\"u}rek, Asli},
  title        = {{Video-mediated multimodal referential communication dataset}},
  year         = {2025},
  month        = apr,
  publisher    = {MPI for Psycholinguistics Archive},
  howpublished = {Collection},
  url          = {https://hdl.handle.net/1839/9012893a-bd5b-4631-8ba3-b7bee4b3a7a1},
  note         = {Accessed: 2026-04-29}
}

@article{alibali2001-gesture,
title = {Effects of Visibility between Speaker and Listener on Gesture Production: Some Gestures Are Meant to Be Seen},
journal = {Journal of Memory and Language},
volume = {44},
number = {2},
pages = {169-188},
year = {2001},
issn = {0749-596X},
doi = {https://doi.org/10.1006/jmla.2000.2752},
url = {https://www.sciencedirect.com/science/article/pii/S0749596X00927529},
author = {Martha W. Alibali and Dana C. Heath and Heather J. Myers},
}

@article{bavelas2002-gesture,
  title={An experimental study of when and how speakers use gestures to communicate},
  author={Bavelas, Janet and Kenwood, Christine and Johnson, Trudy and Phillips, Bruce},
  journal={Gesture},
  volume={2},
  number={1},
  pages={1--17},
  year={2002},
  publisher={John Benjamins}
}

@article{beattie1999-iconic,
  author  = {Beattie, Geoffrey and Shovelton, Heather},
  title   = {Do Iconic Hand Gestures Really Contribute Anything to the
             Semantic Information Conveyed by Speech?
             An Experimental Investigation},
  journal = {Semiotica},
  year    = {1999},
  volume  = {123},
  number  = {1--2},
  pages   = {1--30},
  doi     = {10.1515/semi.1999.123.1-2.1}
}

@article{brennan1996-conceptual,
  author  = {Brennan, Susan E. and Clark, Herbert H.},
  title   = {Conceptual Pacts and Lexical Choice in Conversation},
  journal = {Journal of Experimental Psychology: Learning, Memory, and Cognition},
  volume  = {22},
  number  = {6},
  pages   = {1482--1493},
  year    = {1996},
  doi     = {10.1037/0278-7393.22.6.1482}
}

@article{burkner_brms_2017,
	title = {brms: {An} {R} {Package} for {Bayesian} {Multilevel} {Models} {Using} {Stan}},
	volume = {80},
	doi = {10.18637/jss.v080.i01},
	number = {1},
	journal = {Journal of Statistical Software},
	author = {Bürkner, Paul-Christian},
	year = {2017},
	pages = {1--28},
}

@inProceedings{chen2021-yourefit,
    title={{Y}ou{R}ef{I}t: Embodied Reference Understanding with Language and Gesture},
    author = {Chen, Yixin and Li, Qing and Kong, Deqian and Kei, Yik Lun and Zhu, Song-Chun and Gao, Tao and Zhu, Yixin and Huang, Siyuan},
    booktitle={The IEEE International Conference on Computer Vision (ICCV)},
    year={2021}
}

@article{clark1986-referring,
    title = {Referring as a collaborative process},
    journal = {Cognition},
    volume = {22},
    number = {1},
    pages = {1-39},
    year = {1986},
    issn = {0010-0277},
    doi = {https://doi.org/10.1016/0010-0277(86)90010-7},
    url = {https://www.sciencedirect.com/science/article/pii/0010027786900107},
    author = {Herbert H. Clark and Deanna Wilkes-Gibbs},
}

@INPROCEEDINGS{devries2017-guesswhat,
  author={De Vries, Harm and Strub, Florian and Chandar, Sarath and Pietquin, Olivier and Larochelle, Hugo and Courville, Aaron},
  booktitle={2017 IEEE Conference on Computer Vision and Pattern Recognition (CVPR)}, 
  title={GuessWhat?! {V}isual Object Discovery through Multi-modal Dialogue}, 
  year={2017},
  volume={},
  number={},
  pages={4466-4475},
  doi={10.1109/CVPR.2017.475}
  }

@article{eijk2022-cabb,
    title = {The {CABB} dataset: A multimodal corpus of communicative interactions for behavioural and neural analyses},
    journal = {NeuroImage},
    volume = {264},
    pages = {119734},
    year = {2022},
    issn = {1053-8119},
    doi = {https://doi.org/10.1016/j.neuroimage.2022.119734},
    url = {https://www.sciencedirect.com/science/article/pii/S1053811922008552},
    author = {Lotte Eijk and Marlou Rasenberg and Flavia Arnese and Mark Blokpoel and Mark Dingemanse and Christian F. Doeller and Mirjam Ernestus and Judith Holler and Branka Milivojevic and Asli Özyürek and Wim Pouw and Iris {van Rooij} and Herbert Schriefers and Ivan Toni and James Trujillo and Sara Bögels},
}

@inproceedings{ghaleb2025-cospeech,
    title = "{I} see what you mean: Co-Speech Gestures for Reference Resolution in Multimodal Dialogue",
    author = "Ghaleb, Esam  and
      Khaertdinov, Bulat  and
      Ozyurek, Asli  and
      Fern{\'a}ndez, Raquel",
    editor = "Che, Wanxiang  and
      Nabende, Joyce  and
      Shutova, Ekaterina  and
      Pilehvar, Mohammad Taher",
    booktitle = "Findings of the Association for Computational Linguistics: ACL 2025",
    month = jul,
    year = "2025",
    address = "Vienna, Austria",
    publisher = "Association for Computational Linguistics",
    url = "https://aclanthology.org/2025.findings-acl.682/",
    doi = "10.18653/v1/2025.findings-acl.682",
    pages = "13191--13206",
    ISBN = "979-8-89176-256-5"
}

@incollection{grice1975-logic,
  author    = {Grice, H. Paul},
  title     = {Logic and Conversation},
  booktitle = {Syntax and Semantics, Volume 3: Speech Acts},
  editor    = {Cole, Peter and Morgan, Jerry L.},
  pages     = {41--58},
  publisher = {Academic Press},
  year      = {1975}
}

@inproceedings{haber2019-photobook,
    title = "The {P}hoto{B}ook Dataset: Building Common Ground through Visually-Grounded Dialogue",
    author = {Haber, Janosch  and
      Baumg{\"a}rtner, Tim  and
      Takmaz, Ece  and
      Gelderloos, Lieke  and
      Bruni, Elia  and
      Fern{\'a}ndez, Raquel},
    editor = "Korhonen, Anna  and
      Traum, David  and
      M{\`a}rquez, Llu{\'i}s",
    booktitle = "Proceedings of the 57th Annual Meeting of the Association for Computational Linguistics",
    month = jul,
    year = "2019",
    address = "Florence, Italy",
    publisher = "Association for Computational Linguistics",
    url = "https://aclanthology.org/P19-1184/",
    doi = "10.18653/v1/P19-1184",
    pages = "1895--1910"
}

@article{hanna2003-common,
    title = {The effects of common ground and perspective on domains of referential interpretation},
    journal = {Journal of Memory and Language},
    volume = {49},
    number = {1},
    pages = {43-61},
    year = {2003},
    issn = {0749-596X},
    doi = {https://doi.org/10.1016/S0749-596X(03)00022-6},
    url = {https://www.sciencedirect.com/science/article/pii/S0749596X03000226},
    author = {Joy E Hanna and Michael K Tanenhaus and John C Trueswell},
}

@ARTICLE{hinnell2020-gesture,
    AUTHOR={Hinnell, Jennifer  and Parrill, Fey },
    TITLE={Gesture Influences Resolution of Ambiguous Statements of Neutral and Moral Preferences},
    JOURNAL={Frontiers in Psychology},
    VOLUME={Volume 11 - 2020},
    YEAR={2020},
    URL={https://www.frontiersin.org/journals/psychology/articles/10.3389/fpsyg.2020.587129},
    DOI={10.3389/fpsyg.2020.587129},
    ISSN={1664-1078},
}

@article{holle2007-iconic,
  author  = {Holle, Henning and Gunter, Thomas C.},
  title   = {The Role of Iconic Gestures in Speech Disambiguation:
             {ERP} Evidence},
  journal = {Journal of Cognitive Neuroscience},
  year    = {2007},
  volume  = {19},
  number  = {7},
  pages   = {1175--1192},
  doi     = {10.1162/jocn.2007.19.7.1175}
}

@article{holler_effect_2007,
	title = {The {Effect} of {Common} {Ground} on {How} {Speakers} {Use} {Gesture} and {Speech} to {Represent} {Size} {Information}},
	volume = {26},
	issn = {0261-927X},
	doi = {10.1177/0261927X06296428},
	number = {1},
	urldate = {2026-07-29},
	journal = {Journal of Language and Social Psychology},
	publisher = {SAGE Publications Inc},
	author = {Holler, Judith and Stevens, Rachel},
	year = {2007},
	pages = {4--27},
}

@inproceedings{hu2022-lora,
    title={Lo{RA}: Low-Rank Adaptation of Large Language Models},
    author={Edward J Hu and yelong shen and Phillip Wallis and Zeyuan Allen-Zhu and Yuanzhi Li and Shean Wang and Lu Wang and Weizhu Chen},
    booktitle={International Conference on Learning Representations},
    year={2022},
    url={https://openreview.net/forum?id=nZeVKeeFYf9}
}

@inproceedings{islam2022-caesar,
 author = {Islam, Md Mofijul and Mirzaiee, Reza and Gladstone, Alexi and Green, Haley and Iqbal, Tariq},
 booktitle = {Advances in Neural Information Processing Systems},
 editor = {S. Koyejo and S. Mohamed and A. Agarwal and D. Belgrave and K. Cho and A. Oh},
 pages = {21001--21015},
 publisher = {Curran Associates, Inc.},
 title = {{CAESAR}: An Embodied Simulator for Generating Multimodal Referring Expression Datasets},
 url = {https://proceedings.neurips.cc/paper_files/paper/2022/file/844f722dbbcb27933ff5baf58a1f00c8-Paper-Datasets_and_Benchmarks.pdf},
 volume = {35},
 year = {2022}
}

@article{kita2003-semantic,
    title = {What does cross-linguistic variation in semantic coordination of speech and gesture reveal?: Evidence for an interface representation of spatial thinking and speaking},
    journal = {Journal of Memory and Language},
    volume = {48},
    number = {1},
    pages = {16-32},
    year = {2003},
    issn = {0749-596X},
    doi = {https://doi.org/10.1016/S0749-596X(02)00505-3},
    url = {https://www.sciencedirect.com/science/article/pii/S0749596X02005053},
    author = {Kita, Sotaro and {\"O}zy{\"u}rek, Asl{\i}},
}

@inproceedings{li2023-touchline,
    title={Understanding Embodied Reference with Touch-Line Transformer},
    author={Yang Li and Xiaoxue Chen and Hao Zhao and Jiangtao Gong and Guyue Zhou and Federico Rossano and Yixin Zhu},
    booktitle={The Eleventh International Conference on Learning Representations },
    year={2023},
    url={https://openreview.net/forum?id=ugA1HX69sf}
}

@inproceedings{li2025-unisign,
title={Uni-{S}ign: Toward Unified Sign Language Understanding at Scale},
author={Zecheng Li and Wengang Zhou and Weichao Zhao and Kepeng Wu and Hezhen Hu and Houqiang Li},
booktitle={The Thirteenth International Conference on Learning Representations},
year={2025},
url={https://openreview.net/forum?id=0Xt7uT04cQ}
}

@inproceedings{liu2023-ambiguity,
    title = "We{'}re Afraid Language Models Aren{'}t Modeling Ambiguity",
    author = "Liu, Alisa  and
      Wu, Zhaofeng  and
      Michael, Julian  and
      Suhr, Alane  and
      West, Peter  and
      Koller, Alexander  and
      Swayamdipta, Swabha  and
      Smith, Noah  and
      Choi, Yejin",
    editor = "Bouamor, Houda  and
      Pino, Juan  and
      Bali, Kalika",
    booktitle = "Proceedings of the 2023 Conference on Empirical Methods in Natural Language Processing",
    month = dec,
    year = "2023",
    address = "Singapore",
    publisher = "Association for Computational Linguistics",
    url = "https://aclanthology.org/2023.emnlp-main.51/",
    doi = "10.18653/v1/2023.emnlp-main.51",
    pages = "790--807"
}

@inproceedings{lugaresi2019-mediapipe,
title	= {{M}edia{P}ipe: A Framework for Perceiving and Processing Reality},
author	= {Camillo Lugaresi and Jiuqiang Tang and Hadon Nash and Chris McClanahan and Esha Uboweja and Michael Hays and Fan Zhang and Chuo-Ling Chang and Ming Yong and Juhyun Lee and Wan-Teh Chang and Wei Hua and Manfred Georg and Matthias Grundmann},
year	= {2019},
URL	= {https://mixedreality.cs.cornell.edu/s/NewTitle_May1_MediaPipe_CVPR_CV4ARVR_Workshop_2019.pdf},
booktitle	= {Third Workshop on Computer Vision for AR/VR at IEEE Computer Vision and Pattern Recognition (CVPR) 2019}}

@inproceedings{ma2025-pragmatics,
    title = "Pragmatics in the Era of Large Language Models: A Survey on Datasets, Evaluation, Opportunities and Challenges",
    author = "Ma, Bolei  and
      Li, Yuting  and
      Zhou, Wei  and
      Gong, Ziwei  and
      Liu, Yang Janet  and
      Jasinskaja, Katja  and
      Friedrich, Annemarie  and
      Hirschberg, Julia  and
      Kreuter, Frauke  and
      Plank, Barbara",
    editor = "Che, Wanxiang  and
      Nabende, Joyce  and
      Shutova, Ekaterina  and
      Pilehvar, Mohammad Taher",
    booktitle = "Proceedings of the 63rd Annual Meeting of the Association for Computational Linguistics (Volume 1: Long Papers)",
    month = jul,
    year = "2025",
    address = "Vienna, Austria",
    publisher = "Association for Computational Linguistics",
    url = "https://aclanthology.org/2025.acl-long.425/",
    doi = "10.18653/v1/2025.acl-long.425",
    pages = "8679--8696",
    ISBN = "979-8-89176-251-0"
}

@article{makowski_bayestestr_2019,
	title = {{bayestestR}: {Describing} {Effects} and their {Uncertainty}, {Existence} and {Significance} within the {Bayesian} {Framework}},
	volume = {4},
	doi = {10.21105/joss.01541},
	number = {40},
	journal = {Journal of Open Source Software},
	publisher = {The Open Journal},
	author = {Makowski, Dominique and Ben-Shachar, Mattan S. and Lüdecke, Daniel},
	year = {2019},
	pages = {1541},
}

@book{mcneill1992-hand,
  author    = {McNeill, David},
  title     = {Hand and Mind: What Gestures Reveal about Thought},
  year      = {1992},
  publisher = {University of Chicago Press},
  address   = {Chicago}
}

@misc{mmpose2020,
    title={Open{MM}Lab Pose Estimation Toolbox and Benchmark},
    author={{MMPose Contributors}},
    howpublished = {\url{https://github.com/open-mmlab/mmpose}},
    year={2020}
}

@article{mol2011-seeing,
    author = {Mol, Lisette and Krahmer, Emiel and Maes, Alfons and Swerts, Marc},
    title = {Seeing and Being Seen: The Effects on Gesture Production},
    journal = {Journal of Computer-Mediated Communication},
    volume = {17},
    number = {1},
    pages = {77-100},
    year = {2011},
    month = {10},
    issn = {1083-6101},
    doi = {10.1111/j.1083-6101.2011.01558.x},
    url = {https://doi.org/10.1111/j.1083-6101.2011.01558.x},
    eprint = {https://academic.oup.com/jcmc/article-pdf/17/1/77/19442100/jjcmcom0077.pdf},
}

@inproceedings{yan2018-stgcn,
author = {Yan, Sijie and Xiong, Yuanjun and Lin, Dahua},
title = {Spatial temporal graph convolutional networks for skeleton-based action recognition},
year = {2018},
isbn = {978-1-57735-800-8},
publisher = {AAAI Press},
articleno = {912},
numpages = {9},
location = {New Orleans, Louisiana, USA},
series = {AAAI'18/IAAI'18/EAAI'18}
}

@InProceedings{radford2021-clip,
  title = 	 {Learning Transferable Visual Models From Natural Language Supervision},
  author =       {Radford, Alec and Kim, Jong Wook and Hallacy, Chris and Ramesh, Aditya and Goh, Gabriel and Agarwal, Sandhini and Sastry, Girish and Askell, Amanda and Mishkin, Pamela and Clark, Jack and Krueger, Gretchen and Sutskever, Ilya},
  booktitle = 	 {Proceedings of the 38th International Conference on Machine Learning},
  pages = 	 {8748--8763},
  year = 	 {2021},
  editor = 	 {Meila, Marina and Zhang, Tong},
  volume = 	 {139},
  series = 	 {Proceedings of Machine Learning Research},
  month = 	 {18--24 Jul},
  publisher =    {PMLR},
  url = 	 {https://proceedings.mlr.press/v139/radford21a.html},
}

@article{sedivy1999-contextual,
title = {Achieving incremental semantic interpretation through contextual representation},
journal = {Cognition},
volume = {71},
number = {2},
pages = {109-147},
year = {1999},
issn = {0010-0277},
doi = {https://doi.org/10.1016/S0010-0277(99)00025-6},
url = {https://www.sciencedirect.com/science/article/pii/S0010027799000256},
author = {Julie C. Sedivy and Michael {K. Tanenhaus} and Craig G. Chambers and Gregory N. Carlson},
}

@inproceedings{takmaz2020-refer,
    title = "{R}efer, {R}euse, {R}educe: {G}enerating {S}ubsequent {R}eferences in {V}isual and {C}onversational {C}ontexts",
    author = "Takmaz, Ece  and
      Giulianelli, Mario  and
      Pezzelle, Sandro  and
      Sinclair, Arabella  and
      Fern{\'a}ndez, Raquel",
    editor = "Webber, Bonnie  and
      Cohn, Trevor  and
      He, Yulan  and
      Liu, Yang",
    booktitle = "Proceedings of the 2020 Conference on Empirical Methods in Natural Language Processing (EMNLP)",
    month = nov,
    year = "2020",
    address = "Online",
    publisher = "Association for Computational Linguistics",
    url = "https://aclanthology.org/2020.emnlp-main.353/",
    doi = "10.18653/v1/2020.emnlp-main.353",
    pages = "4350--4368",
}

@article{turner_perspective_2026,
	title = {Perspective {Taking} {Increases} {Representational} {Gesture} {Production} in {Videoconferencing} {Settings}},
	volume = {50},
	issn = {1573-3653},
	doi = {10.1007/s10919-026-00502-w},
	number = {2},
	journal = {Journal of Nonverbal Behavior},
	author = {Turner, Sydney and Young, Sylvia E. and Kita, Sotaro and Morett, Laura M.},
	year = {2026},
	pages = {195--209},
}

@inproceedings{udagawa2019-onecommon,
  title={A natural language corpus of common grounding under continuous and partially-observable context},
  author={Udagawa, Takuma and Aizawa, Akiko},
  booktitle={Proceedings of the AAAI Conference on Artificial Intelligence},
  volume={33},
  number={01},
  pages={7120--7127},
  year={2019}
}

@INPROCEEDINGS{zhang2024-dstformer,
  author={Zhang, Shaokun and Li, Xinde and Hu, Chuanfei and Xu, Jianping and Liu, Huaping},
  booktitle={2024 International Conference on Advanced Robotics and Mechatronics (ICARM)}, 
  title={{DSTF}ormer: 3D {H}uman Pose Estimation with a Dual-scale Spatial and Temporal Transformer Network}, 
  year={2024},
  volume={},
  number={},
  pages={484-489},
  doi={10.1109/ICARM62033.2024.10715863}
}

@misc{qwen3,
  title={Qwen3-{VL} Technical Report}, 
  author={Shuai Bai and Yuxuan Cai and Ruizhe Chen and Keqin Chen and Xionghui Chen and Zesen Cheng and Lianghao Deng and Wei Ding and Chang Gao and Chunjiang Ge and Wenbin Ge and Zhifang Guo and Qidong Huang and Jie Huang and Fei Huang and Binyuan Hui and Shutong Jiang and Zhaohai Li and Mingsheng Li and Mei Li and Kaixin Li and Zicheng Lin and Junyang Lin and Xuejing Liu and Jiawei Liu and Chenglong Liu and Yang Liu and Dayiheng Liu and Shixuan Liu and Dunjie Lu and Ruilin Luo and Chenxu Lv and Rui Men and Lingchen Meng and Xuancheng Ren and Xingzhang Ren and Sibo Song and Yuchong Sun and Jun Tang and Jianhong Tu and Jianqiang Wan and Peng Wang and Pengfei Wang and Qiuyue Wang and Yuxuan Wang and Tianbao Xie and Yiheng Xu and Haiyang Xu and Jin Xu and Zhibo Yang and Mingkun Yang and Jianxin Yang and An Yang and Bowen Yu and Fei Zhang and Hang Zhang and Xi Zhang and Bo Zheng and Humen Zhong and Jingren Zhou and Fan Zhou and Jing Zhou and Yuanzhi Zhu and Ke Zhu},
  year={2025},
  eprint={2511.21631},
  archivePrefix={arXiv},
  primaryClass={cs.CV},
  url={https://arxiv.org/abs/2511.21631}, 
}

@article{rasenberg2022primacy,
  title={The primacy of multimodal alignment in converging on shared symbols for novel referents},
  author={Rasenberg, Marlou and {\"O}zy{\"u}rek, Asli and B{\"o}gels, Sara and Dingemanse, Mark},
  journal={Discourse Processes},
  volume={59},
  number={3},
  pages={209--236},
  year={2022},
  publisher={Taylor \& Francis}
}

@inproceedings{cheng2020decoupling,
  title={Decoupling gcn with dropgraph module for skeleton-based action recognition},
  author={Cheng, Ke and Zhang, Yifan and Cao, Congqi and Shi, Lei and Cheng, Jian and Lu, Hanqing},
  booktitle={European conference on computer vision},
  pages={536--553},
  year={2020},
  organization={Springer}
}

@inproceedings{jiang2021skeleton,
  title={Skeleton aware multi-modal sign language recognition},
  author={Jiang, Songyao and Sun, Bin and Wang, Lichen and Bai, Yue and Li, Kunpeng and Fu, Yun},
  booktitle={2021 IEEE/CVF Conference on Computer Vision and Pattern Recognition Workshops (CVPRW)},
  pages={3408--3418},
  year={2021},
  organization={IEEE}
}

@inproceedings{tiedemann2020opus,
  title={{OPUS}-{MT}--{B}uilding open translation services for the World},
  author={Tiedemann, J{\"o}rg and Thottingal, Santhosh},
  booktitle={Annual Conference of the European Association for Machine Translation},
  pages={479--480},
  year={2020},
  organization={European Association for Machine Translation}
}

\appendix
\section{Palette of all Fribbles}\label{appendix:all_fribbles}
\begin{figure*}
    \centering
    \includegraphics[width=0.9\linewidth]{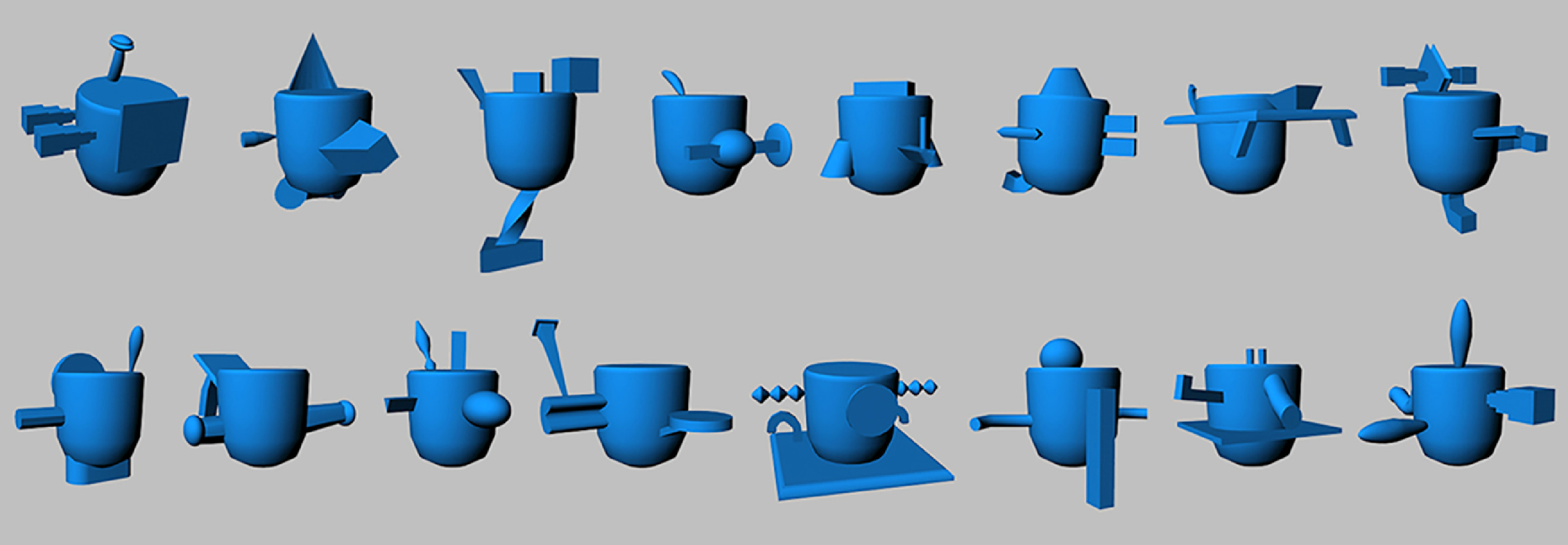}
    \caption{Palette of all 16 Fribbles used in the original CABB studies. Image source: \citet{eijk2022-cabb}.}
    \label{fig:all_fribbles}
\end{figure*}

Figure~\ref{fig:all_fribbles} displays the complete set of 16 Fribbles shown to participants in the original face-to-face CABB study \citep{eijk2022-cabb} as well as the study on video-mediated interaction \citep{akamine2025-cabb-videomediated, akamine2025-dataset}.
We use the same set of Fribbles in our study of models.

\section{General-purpose vision--language model baseline}\label{appendix:qwen}
We evaluate Qwen3-VL \citep{qwen3} as a general-purpose baseline for multimodal reference resolution.
Unlike the controlled models in the main experiments, which use extracted skeletal representations, Qwen3-VL receives gesture video frames directly.
This comparison tests whether a pretrained vision--language model can recover the intended referent from the transcript, the visible gesture, or their combination.

\subsection{Task formulation}
We formulate reference resolution as a 16-way multiple-choice task.
For each example, the 16 Fribbles are arranged in a $4 \times 4$ grid on a white background, with each cell labelled by a letter from A to P.
The arrangement is randomly shuffled for every example, so that the target Fribble does not have a fixed position or letter across samples.
The model must predict the letter assigned to the target in the current grid.
This design requires the model to ground the referring information in the displayed candidates rather than memorising an association between a Fribble and a particular label.

We evaluate Qwen3-VL under transcript-only, gesture-only, and multimodal input conditions.
Gesture input consists of eight frames samples from the annotated gesture interval.
Transcript input consists of the aligned English translation of the overlapping utterance.
In the multimodal condition, the model receives both the transcript and the gesture frames.
In all conditions, Qwen3-VL is shown the candidate grid.

\subsubsection{Transcript-only prompt template}
\texttt{You see the transcript of what the speaker said during a gesture and a grid of labelled objects (A–P).\\
Question: Which letter labels the object being referred to? Reply with a single letter.\\\\
Answer:}

\subsubsection{Gesture-only prompt template}
\texttt{You see a short clip of the speaker gesturing and a grid of labelled objects (A–P).\\
Question: Which letter labels the object being referred to? Reply with a single letter.\\\\
Answer:}

\subsubsection{Multimodal prompt template}
\texttt{You see a short clip of the speaker gesturing, the transcript of what was said, and a grid of labelled objects (A–P).\\
Question: Which letter labels the object being referred to? Reply with a single letter.\\\\
Answer:}

\subsection{Untuned inference}
We first evaluate Qwen3-VL without task-specific fine-tuning.
Inference consists of a single forward pass followed by an argmax over the 16 A--P token logits.
No temperature scaling, top-$k$, or nucleus sampling was used.
With the 4B model, transcript-only input achieves 12.85\% accuracy, while gesture-only input achieves 6.12\%.
Combining the two modalities results in 12.80\%, providing no improvement over the transcript alone.

To examine whether this result is primarily due to model scale, we also evaluate the 32B variant.
Scaling improves transcript-only accuracy to 19.12\%, but gesture-only performance remains low at 6.29\%.
Multimodal accuracy reaches 19.10\%, again closely matching the transcript-only results.
Thus, under untuned inference, Qwen3-VL derives most of its predictive signal from the transcript and does not appear to make effective use of the gesture frames.

\subsection{LoRA fine-tuning}
We next fine-tune Qwen3-VL-4B under LoRA \citep{hu2022-lora} under four training configurations.
All configurations use the same shuffled candidate-grid formulation and are evaluated using five-fold cross-validation.

\subsubsection{Technical details}
LoRA adapters were applied to the query and value projection layers with rank 8, scaling factor 16, and dropout 0.05.
This resulted in 2.95 million trainable parameters, corresponding to 0.066\% of the 4.44-billion-parameter model.
Training used bfloat16 precision without weight quantisation and employed gradient checkpointing.
We optimised the adapters using AdamW with a constant learning rate of $2\times 10^{-4}$, weight decay of 0.01, and no learning-rate scheduler or warm-up.
Each model was trained for eight epochs.

We performed 5-fold cross-validation.
In each cross-validation iteration, one bucket was used as the test set, the subsequent buckets as the development set, and the remaining three buckets as the training set.
Training minimised 16-class cross-entropy over the model's logits for the single-token answers A--P at the final prompt position.
Best performing checkpoints were selected based on development-set accuracy.

\subsubsection{Results}
In the \textbf{gesture-only condition}, the model receives eight gesture frames and the Fribble grid, without a transcript.
Fine-tuning raises accuracy to $16.89\% \pm 1.77$, indicating that the model can learn some referential information from the gesture videos.

In the \textbf{transcript-only condition}, the model receives the translated utterance and candidate grid.
Because several annotated gestures may overlap the same utterance, retaining every gesture-centred instance would duplicate identical transcript inputs.
We therefore deduplicate the transcripts in this condition.
The resulting model achieves $52.54\% \pm 3.66$ accuracy.

For \textbf{multimodal fine-tuning}, we consider two sampling strategies.
First, we retain one randomly selected gesture for each gesture-overlapping transcript, controlling for the repeated occurrence of transcripts that overlap multiple gestures.
This configuration achieves $52.42\% \pm 3.34$.
Second, we retain all gesture--transcript pairings, allowing the same transcript to occur with each overlapping gesture.
This configuration achieves $52.62\% \pm 4.01$.

\subsection{Interpretation}
Task-specific fine-tuning substantially improves Qwen3-VL, particularly when transcripts are available.
However, neither multimodal sampling strategy produces a clear improvement over the transcript-only model.
The similarity between the two multimodal results also suggests that repeated transcript occurrences do not explain the absence of a fusion gain.

This negative result does not necessarily imply that gesture lacks complementary referential information.
In Qwen3-VL, gesture perception, temporal representation, multimodal integration, and referent selection are handled within a single general-purpose architecture.
It is therefore unclear whether the model fails because the gesture signal is uninformative or because it does not extract and integrate that signal effectively.
The controlled skeletal models in the main experiments separate these components and allow us to test more directly whether gesture contains information beyond the transcript.

\section{Posterior summary statistics and data visualization}
\label{app:posterior}

\begin{figure}[h!]
    \centering
    \includegraphics[width=\linewidth]{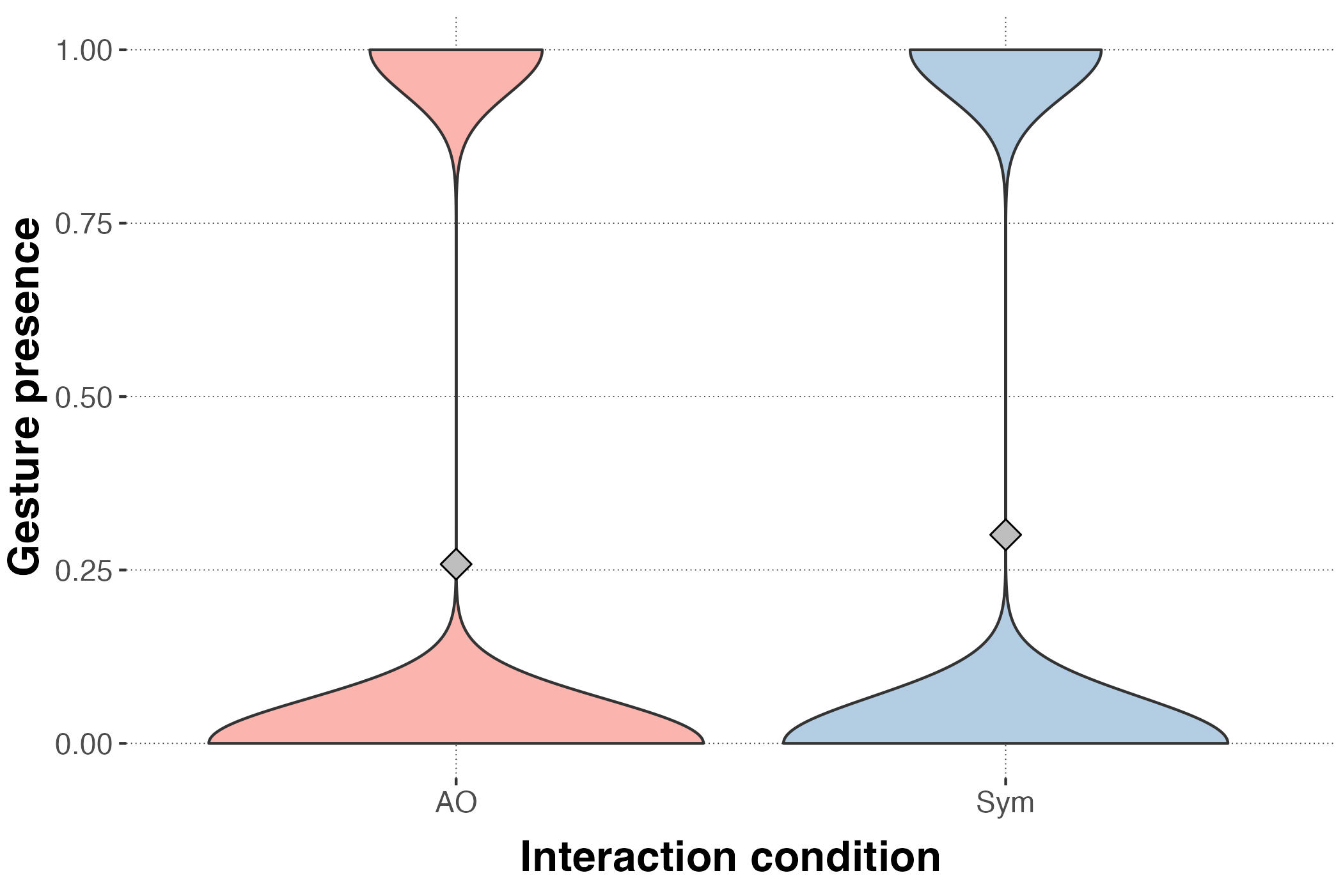}
    \caption{Human use of gestures by interaction condition in the dataset.}
    \label{fig:human_gesture_cond}
\end{figure}

\begin{figure}[h!]
    \centering
    \includegraphics[width=\linewidth]{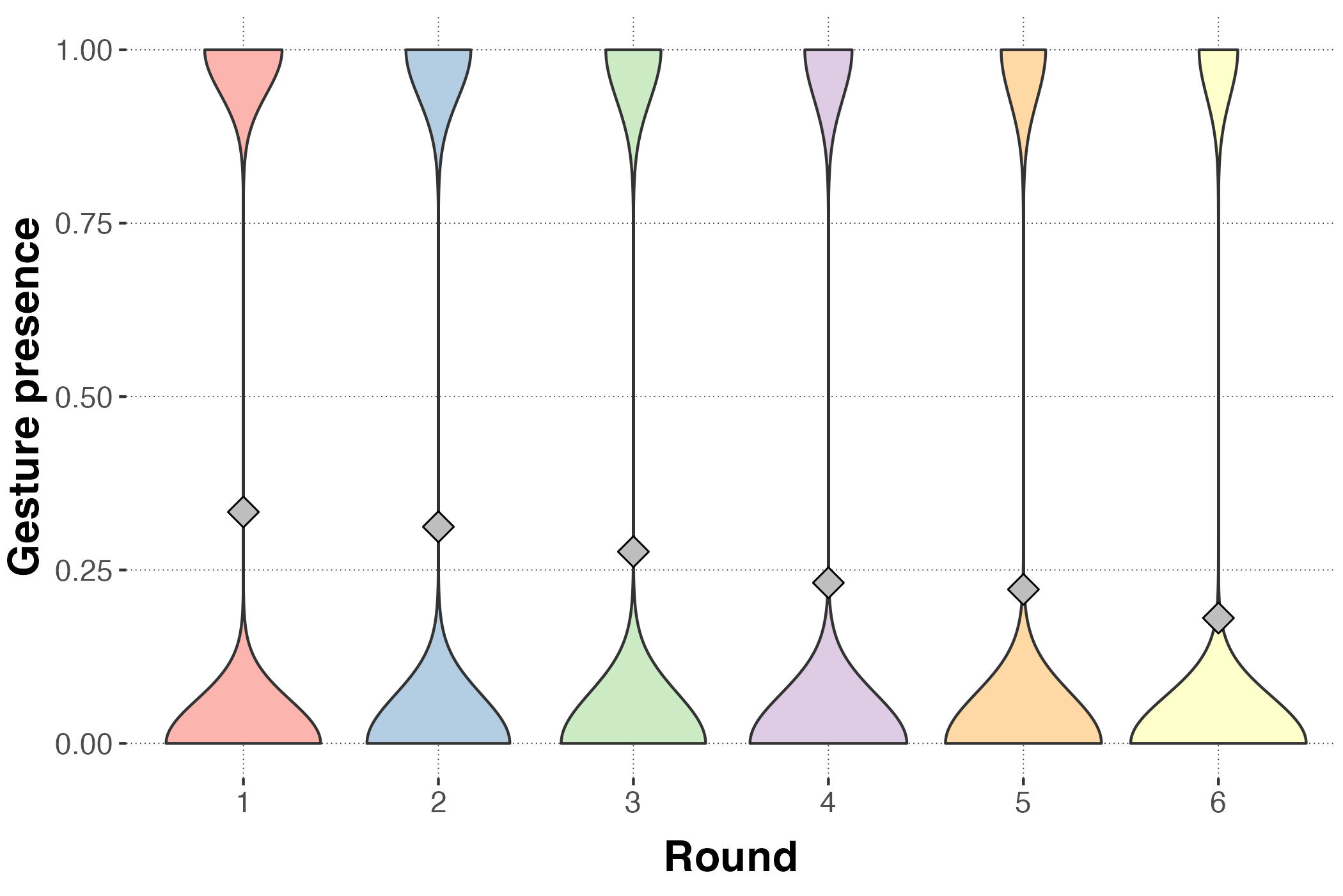}
    \caption{Distribution of gestures across rounds in the dataset of human interactions.}
    \label{fig:round_gesture_human}
\end{figure}

We provide summary statistics and additional data visualisations for the Bayesian logistic regression models run for the effects of interlocutor visibility and entrainment across rounds on human interactions and model performance. We defined gesture presence as a speech segment where gesture is present (1) or not (0). This variable is binary coded. We thus use a bernoulli model with logit link function to predict it. We use hierarchical models to capture group-level effects per participant pair. As priors, we use a normal distribution centered on -1 with a standard deviation of 0.5 for the intercept, a direction-agnostic normal distribution centered on 0 with a standard deviation of 0.1 for the population-level effects and an LKJ prior for correlations. Interlocutor visibility results are presented in Table~\ref{tab:gesture_present_visibility_human} and Figure~\ref{fig:human_gesture_cond} (human data) and Table~\ref{tab:posterior_summary_comm} (models).
Entrainment results are presented in Table~\ref{tab:gesture_present_round_human} and Figure~\ref{fig:round_gesture_human} (human data) and Table~\ref{tab:gesture_present_round_model} (models). We report Means and Credible Intervals measured as the Highest Density Interval for all model coefficients. The Rhat values and Effective Sample Sizes (ESS) are reported to assess model convergence and indicate excellent model convergence (<1.01) and enough posterior samples drawn from the distribution (ESS > 1000). We additionally report results from a Bayesian hypothesis test. The probability of direction (pd) indicates how much of the posterior is of the same sign and can be interpreted alongside the frequentist p-value, with 97.5\% corresponding to p<.05. We used the brms and bayestestR packages for statistical analyses~\citep{burkner_brms_2017, makowski_bayestestr_2019}. 

\begin{table*}
\caption{\label{tab:gesture_present_visibility_human}Posterior summary statistics for the visibility condition-based model of gesture presence in human data}
\resizebox{\textwidth}{!}{%
\centering
\begin{tabular}[t]{cccccc}
\toprule
Parameter & Mean & CrI & pd & Rhat & ESS\\
\midrule
b\_Intercept & -1.11 & {}[-1.37, -0.831] & 100\% & 1 & 2148.07\\
b\_conditionSym & 0.03 & {}[-0.156, 0.212] & 62\% & 1 & 11677.85\\
\bottomrule
\end{tabular}
}
\end{table*}

\begin{table*}
\caption{\label{tab:posterior_summary_comm}Posterior summary statistics for the visibility condition-based model of model accuracies}
\resizebox{\textwidth}{!}{%
\centering
\begin{tabular}[t]{lccccccc}
\toprule
Parameter & Mean & CrI & pd & Rhat & ESS\\
\midrule
b\_Intercept & -1.52 & {}[-1.617, -1.425] & 100.00\%& 1 & 1519.81\\
b\_ConditionSym & 0.19 & {}[0.054, 0.315] & 99.75\% & 1 & 1504.34\\
b\_input\_modespeech & 1.36 & {}[1.231, 1.476] & 100.00\% & 1 & 1669.59\\
b\_input\_modeboth & 1.49 & {}[1.362, 1.608] & 100.00\% & 1 & 1533.49\\
b\_ConditionSym:input\_modespeech & -0.22 & {}[-0.383, -0.056] & 99.50\% & 1 & 1612.13\\
b\_ConditionSym:input\_modeboth & -0.24 & {}[-0.403, -0.074] & 99.88\% & 1 & 1484.87\\
\bottomrule
\end{tabular}
}
\end{table*}

\begin{table*}
\caption{\label{tab:gesture_present_round_human}Posterior summary statistics for the round-based model of gesture-presence in the human data}
\resizebox{\textwidth}{!}{%
\centering
\begin{tabular}[t]{cccccc}
\toprule
Parameter & Mean & CrI & pd & Rhat & ESS\\
\midrule
b\_Intercept & -0.55 & {}[-0.803, -0.289] & 100\% & 1 & 2274.25\\
b\_round & -0.21 & {}[-0.281, -0.14] & 100\% & 1 & 2979.91\\
\bottomrule
\end{tabular}
}
\end{table*}

\begin{table*}
\caption{\label{tab:gesture_present_round_model}Posterior summary statistics for the round-based model of accuracies}
\resizebox{\textwidth}{!}{%
\centering
\begin{tabular}[t]{lcccc}
\toprule
Parameter & Mean & CrI & pd & Dyad CI\\
\midrule
b\_Intercept & -1.51 & {}[-1.629, -1.398] & 100.00\% & {}[-1.740, -1.321]\\
b\_input\_modespeech & 0.91 & {}[0.760, 1.056] & 100.00\% & {}[0.674, 1.166]\\
b\_input\_modeboth & 0.91 & {}[0.760, 1.055] & 100.00\% & {}[0.684, 1.143]\\
b\_round & 0.04 & {}[0.001, 0.084] & 97.76\% & {}[-0.011, 0.093]\\
b\_input\_modespeech:round & 0.09 & {}[0.037, 0.145] & 99.95\% & {}[0.021, 0.174]\\
b\_input\_modeboth:round & 0.12 & {}[0.068, 0.175] & 100.00\% & {}[0.059, 0.195]\\
\bottomrule
\end{tabular}
}
\end{table*}

\section{Technical details of the proposed models}
\label{app:technical}

\paragraph{STGCN Backbone.}
We use a decoupled variant of \stgcn{} \citep{yan2018-stgcn} in the
SL-GCN lineage \citep{cheng2020decoupling, jiang2021skeleton}.

This architecture consists of multiple graph convolution layers. Each unit is a graph convolution followed by attention and a temporal
convolution with kernel size 9.
The stack has ten units (Table~\ref{tab:stgcn_stack}) with two stride-2 units that reduce the temporal resolution to $T/4$.
The output is mean-pooled over joints and then length-masked mean-pooled over valid frames, giving one 256-dimensional vector per window, which we use as gesture representations. 
\begin{table}[t]
\centering
\small
\setlength{\tabcolsep}{4pt}
\begin{tabular}{@{}lccc@{}}
\toprule
Unit & Width & Stride & Resolution \\
\midrule
$l_1$            & 64  & 1 & $T$   \\
$l_2$--$l_4$     & 64  & 1 & $T$   \\
$l_5$            & 128 & 2 & $T/2$ \\
$l_6$--$l_7$     & 128 & 1 & $T/2$ \\
$l_8$            & 256 & 2 & $T/4$ \\
$l_9$--$l_{10}$  & 256 & 1 & $T/4$ \\
\bottomrule
\end{tabular}
\caption{\stgcn{} unit stack. The pooled gesture representation is 256-dimensional.}
\label{tab:stgcn_stack}
\end{table}

\paragraph{Augmentation.}
We apply conservative pose augmentation during training, including mirror permutation using the validated per-layout flip map, small rotations and scalings, and temporal jitter.

\paragraph{Heads and fusion.}
Speech-transcription dimensions are precomputed 768-dimensional frozen CLIP text features and
projected to 256 dimensions by a layer-normalised MLP with GELU.
For \texttt{both}, the speech-anchored gated residual fusion of
Section~\ref{sec:classification_and_alignment_method} combines the two
256-dimensional branches.

The classifier is a two-layer MLP with hidden width 512, GELU, dropout $0.30$, and 17 outputs.

Alignment projectors map a 256-dimensional student into the frozen
768-dimensional CLIP space and are discarded at test time.

\subsection{DSTFormer}
For the benchmarked DSTFormer, the 51-node model has six layers,
width 192, eight heads, and MLP ratio 4; the \unimodel{} model has four layers,
width 128, eight heads, and MLP ratio 2.  Both use learned spatial--temporal
branch fusion and learned temporal pooling, are trained from scratch, and are
excluded from the later modality, condition, and alignment analyses after
ST-GCN is selected.

\paragraph{DSTFormer performance}
DSTFormer's performance was sub-optimal (Table \ref{tab:main_table_dstformer}), perhaps due to the fact that a Transformer-based model would require much more training data. However, STGCN can already exploit the prior pose and its graph efficiently. We therefore proceed with our modelling and experimentation with STGCN. 
\label{sect:dstformer_results}
\begin{table*}[t]
\centering\small
\setlength{\tabcolsep}{3.2pt}
\begin{tabular}{@{}llllccccc@{}}
\toprule
Input & Layout & Backbone & Align. & Top-1 & Macro-F1 & Top-2 & Top-3 & Top-5 \\
\midrule
Gesture & MediaPipe-3D & ST-GCN & --- & 19.5\stdv{1.1} & 18.2\stdv{1.2} & 32.5\stdv{1.8} & 41.6\stdv{1.6} & 55.5\stdv{1.9} \\
Gesture & MMPose-2D & ST-GCN & --- & 20.4\stdv{1.4} & 18.9\stdv{1.3} & 32.6\stdv{2.2} & 42.1\stdv{2.5} & 56.4\stdv{3.3} \\
Gesture & MediaPipe-3D & ST-GCN & P$\rightarrow$V & 20.4\stdv{1.1} & 18.8\stdv{1.0} & 32.5\stdv{1.7} & 41.6\stdv{2.1} & 55.2\stdv{2.3} \\
Gesture & MMPose-2D & ST-GCN & P$\rightarrow$V & 20.1\stdv{1.0} & 18.6\stdv{1.2} & 32.6\stdv{2.2} & 41.5\stdv{2.4} & 55.0\stdv{2.7} \\
Gesture & MediaPipe-3D & DSTFormer & --- & 18.7\stdv{1.5} & 17.5\stdv{1.7} & 30.5\stdv{2.6} & 39.1\stdv{3.7} & 53.3\stdv{4.3} \\
Gesture & MMPose-2D & DSTFormer & --- & 16.3\stdv{1.3} & 14.9\stdv{1.4} & 28.1\stdv{1.7} & 36.9\stdv{2.9} & 51.8\stdv{3.3} \\
\midrule
\textit{chance} & --- & --- & --- & 5.9 & --- & 11.8 & 17.6 & 29.4 \\
\bottomrule
\end{tabular}
\caption{Mean prediction performance over the five test folds. $U_{1:5}$ is the mean top-1--top-5 utility. \dst{} rows are gesture-only and reported for completeness, and they are sub-optimal in comparison to STGCN.}
\label{tab:main_table_dstformer}
\end{table*}

\paragraph{Budget}
STGCN models are lightweight, and the biggest model (which uses \unimodel pose layout) has approximately 13.93 M parameters when it is used with \fstar{}. We list the hyperparameters in Table \ref{tab:optim}. We run the pipeline as a single distributed job over four NVIDIA A100 GPUs using PyTorch DDP. Each cross-validation (with five folds) for a baseline model takes approximately 3 hours. 

\begin{table}[t]
\centering
\small
\setlength{\tabcolsep}{4pt}
\begin{tabular}{@{}lc@{}}
\toprule
Setting & Value \\
\midrule
Optimiser                 & AdamW \\
Backbone / head LR        & $1\times 10^{-4}$ \\
Alignment projector LR    & $3\times 10^{-4}$ \\
Weight decay              & $0.01$ \\
Schedule                  & cosine, $5\%$ linear warm-up \\
Gradient clipping         & $1.0$ \\
Batch (per GPU)           & 16 \\
Gradient accumulation     & 2 \\
Effective batch           & 128 \\
Contrastive pool          & 64 (cross-rank gather) \\
Max epochs                & 256 \\
Early-stopping patience   & 20 \\
Selection metric          & dev macro-F1 \\
Loss                      & class-balanced CE \\
Dropout                   & $0.30$ \\
Alignment warm-up         & 5 epochs \\
Alignment dimension       & 768 (frozen CLIP space) \\
Seed                      & 42 \\
\bottomrule
\end{tabular}
\caption{Optimisation settings, shared across layouts, input modes, and
alignment conditions.}
\label{tab:optim}
\end{table}

\begin{figure}
    \centering
    \includegraphics[width=\linewidth]
    {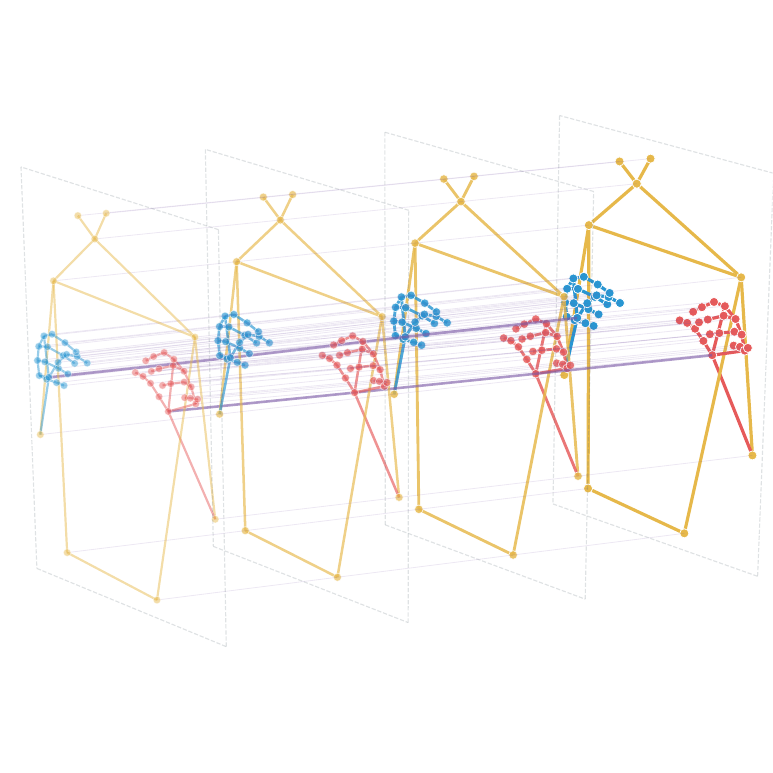}
    \caption{The \mpmodel{} topology: nine upper-body landmarks and both complete 21-landmark hands. Edges follow anatomical connectivity and define both the graph convolution and the bone (parent-offset) channels.}
    \label{fig:mediapipe51_stgcn_3d}
\end{figure}

\section{Dataset usage agreement}
For the subset of the face-to-face CABB dataset \citep{eijk2022-cabb} selected by \citet{rasenberg2022primacy}, data were provided (in part) by Radboud University, Nijmegen, The Netherlands.
For the video-mediated \citep{akamine2025-dataset} versions of CABB, data were provided (in part) by Max Planck Institute for Psycholinguistics, Nijmegen, The Netherlands.
While using these datasets, we have adhered to the Data Use Terms agreed upon when requesting access to the datasets.
\end{document}